\documentclass[11pt,A4paper]{particle}

\usepackage{graphicx}
\usepackage{amsmath}
\usepackage{amsfonts}
\usepackage{dsfont}
\usepackage{subfigure}
\usepackage{hyperref}

\title{Control of rough terrain vehicles using deep reinforcement learning}
\author{Viktor Wiberg$^{*}$, Erik Wallin$^{*}$, Martin Servin$^{*}$, and Tomas Nordfjell$^{\dag}$}

\address{$^{*}$ Umeå University, 
viktor.wiberg@umu.se, erik.wallin@umu.se, martin.servin@umu.se,  http://www.digitalphysics.se/
\and
$^{\dag}$ Swedish University of Agricultural Sciences, tomas.nordfjell@slu.se}

\begin{document}

\begin{abstract}
W
We explore the potential to control terrain vehicles using deep reinforcement in scenarios where human operators and traditional control methods are inadequate.
    This letter presents a controller that perceives, plans, and successfully controls a 16-tonne forestry vehicle with two frame articulation joints, six wheels, and their actively articulated suspensions to traverse rough terrain.
    The carefully shaped reward signal promotes safe, environmental, and efficient driving, which leads to the emergence of unprecedented driving skills.
    We test learned skills in a virtual environment, including terrains reconstructed from high-density laser scans of forest sites.
    The controller displays the ability to handle obstructing obstacles, slopes up to 27$^\circ$, and a variety of natural terrains, all with limited wheel slip, smooth, and upright traversal with intelligent use of the active suspensions.
    The results confirm that deep reinforcement learning has the potential to enhance control of vehicles with complex dynamics and high-dimensional observation data compared to human operators or traditional 
    control methods, especially in rough terrain.
\end{abstract}

\section{INTRODUCTION}
Deep reinforcement learning has recently shown promise for locomotion tasks, but its usefulness to learn control of heavy vehicles in rough terrain is widely unknown.
Conventionally, the design of rough terrain vehicles strives to promote high traversability and be easily operated by humans.
The drivelines involve differentials and bogie suspension that provide ground compliance and reduces the many degrees of freedom, leaving only speed and heading
for the operator to control. An attractive alternative is to use actively articulated suspensions and individual wheel control.
These have the potential to reduce the energy consumption and ground damage, yet increase traversability and tip over stability~\cite{iagnemma2003control, gelin2020concept, wu2017energy, hutter2016, He2019}. 
The concepts have been a reappearing topic in planetary exploration, military, construction, agriculture, and forestry applications, but not yet reached the maturity of practical use~\cite{Gelin2020}.
However, there is reason to believe that the full potential of the vehicles is not being utilized.  
The benefits of active suspension and individual wheel control can only be unlocked
if the many degrees of freedom are controlled at sufficient speed, precision, and
robustness.
Traditional control methods are not well suited to account for the vehicle dynamics and the surrounding environment observed through high-dimensional sensor data, which raises a need for alternatives.

Only in recent years has reinforcement learning (RL) emerged as a candidate
approach for smart control in locomotion applications. Deep learning based control of legged
locomotion demonstrate robustness over a variety of environments and learnt
behaviour not seen before~\cite{lee2020learning}. The success in legged
locomotion indicates the capability of deep RL to learn control of wheeled ground vehicles. However, only
a handful of papers deal with RL applied to wheeled ground
vehicles~\cite{josef2020deep, zhang2018robot}. Local navigation using RL in
rough terrain is addressed in~\cite{josef2020deep} with improved performance
over traditional planning methods. Their application to search and rescue robots
considers safe traversal but discards energy consumption, explicit wheel slip,
and ground damage; important aspects in agriculture and forestry. In addition,
they only use a 3-dimensional, binary control signal. To the best of our
knowledge, RL has not yet been applied to wheeled ground vehicles in rough
terrain with high dimensional, continuous, control signals.

To test the usefulness of deep RL on vehicles in rough terrain, we use physics-based
simulation to develop a controller for a novel concept forwarder, with actively
articulated suspensions and individual control of its six wheels.
Based on a 634-dimensional observation attainable from onboard sensors, we demonstrate learned skills on challenging terrains with steep slopes and obstacles, where performance is measured in terms of our reward signal.
A reward carefully designed to encapsulate safety, energy
consumption, environmental impact, and success of the overall goal; to reach a
specified vehicle pose.
A forestry use case is studied using a forest terrain reconstructed from high-density laser scans, where the controller is assigned a sequence of waypoints along a transport route.
We assess model robustness and domain transferability by varying friction and vehicle load.

\section{BACKGROUND}
Wheeled locomotion in rough terrain involves perceiving the terrain features to make up time and energy-efficient motion plans. Preferably, the motion plans are without risk of getting stuck on obstacles or damaging sensitive parts of the vehicle. Traversing the terrain involves controlling the actuators and making use of sensor data for estimating the current state.
Some wheel slip is inevitable, but excessive slip is associated with ground damage and unnecessary fuel consumption.  Tipping over is a rare but disastrous event, but with higher risk when the vehicle carries a load.

With active suspensions, a vehicle can distribute its load on the wheels to maximise traction or minimise ground pressure, cross otherwise impassable obstacles, and shift its centre of mass to handle inclined terrain.
Individual wheel control can reduce wheel slip and shearing soil surface compared to wheeled and tracked bogies.

We address smart control applied to forestry and the \emph{Xt28 forwarder}
(eXtractor AB). The Xt28 has individual wheel control and actively articulated
suspensions, designed for slopy, rough terrain, and the aim to reduce soil
compaction and shearing. A typical forestry scenario involves an approximate
route, where we assume that a global path planner provides target locations, see
Fig.~\ref{fig:results:LAS-3d}. In cut-to-length logging, the dominating method in
Europe~\cite{lundback2021worldwide}, targets can be manually extracted from the harvester route.
Alternatively, a more general and sophisticated way is to use a trafficability
map~\cite{Guastella2021}. To take into consideration all the
aforementioned rough terrain objectives, coupled with the many control degrees
of freedom of the Xt28 is a challenging task. In this paper, we explore learning
a control policy using reinforcement learning.

\subsection{Reinforcement learning}
Reinforcement learning is a process of interaction between an agent (controller)
and its environment. An environment consists of a state space $\mathcal{S}$,
action space $\mathcal{A}$, transition probabilities $p(s'| s, a)$, and a reward
function $r: \mathcal{S} \times \mathcal{A} \rightarrow \mathbb{R}$. At each
step, the agent selects an action following its policy $a \sim \pi( \cdot, s)$
and current state $s$, and the environment responds with a new state $s'$ and
reward $r = r(s, a)$. The goal of the agent is to maximize the expected future
sum of discounted rewards $\mathbb{E}_\pi \left[ R_t | s_t \right]$, where $R_t
= \sum_{k=0}^\infty \gamma^k r_{t + k + 1}$ is called the return and the
discount factor, $\gamma \in [0, 1]$, values the importance of short-term,
compared to long-term rewards.

In the actor-critic framework, the actor contains the policy, which
in deep RL is modelled as a neural network with parameters $\theta$.
The role of the actor is to sample actions from its policy, $\pi_\theta$,
and adjust its parameters as suggested
by the critic. The critic, or state-value function $V^\pi(s) = \mathbb{E}_\pi
[R_t | s_t]$, evaluates the actor by giving critique to its actions. Most often
the purpose of the state-value function is to compute the \emph{advantage}
$A^\pi(s_t, a_t) = Q^\pi(s_t, a_t) - V^\pi(s_t)$, where the action-value
function is given by $Q^\pi(s_t, a_t) = \mathbb{E}_\pi [R_t | s_t, a_t]$. The
advantage measures the benefit of taking a specific action $a_t$ when in $s_t$
compared to being in that state in general and following policy $\pi_\theta$. It
yields almost the smallest possible variance in policy gradient estimates, but
must be approximated in practice, e.g. using generalize advantage estimate
GAE($\lambda$)~\cite{schulman2015high}.

\subsection{Proximal policy optimization}
Proximal policy optimization (PPO) is an on-policy method which attempts to keep
policy updates close enough to the current policy to improve performance without the risk of
collapse~\cite{schulman2017proximal}. After collecting a batch of samples under
the current policy $\pi_{\theta_k}$, PPO performs minibatch stochastic gradient
decent to find $\theta$ which maximizes the objective~\cite{achiam2018spinning}
\begin{align}
& \mathcal{L}(s, a, \theta_k, \theta) \nonumber \\
    & \qquad = \mathrm{min} \left( \frac{\pi_\theta (a| s)}{\pi_{\theta_k} (a|s)} 
        A^{\pi_{\theta_k}} (s, a),~
        g(\epsilon, A^{\pi_{\theta_k}}(s, a))\right), \\
& g(\epsilon, A) = 
    \begin{cases}
        (1 + \epsilon)A \qquad A \geq 0  \\
        (1 - \epsilon)A \qquad A < 0.
    \end{cases}
\end{align}
The loss motivates policy updates to encourage actions which lead to a positive
advantage and discourage the opposite. To avoid moving too far from the old
policy the objective sets a limit on the policy probability ratio by clipping it
in relation to the advantage, where the clipping range is controlled by the
hyperparameter $\epsilon$. It is common to also include two additional terms in
the loss function. One is an error term on value estimates which is only
necessary if using a network architecture which shares parameters between policy
and value function. The other is an entropy bonus with purpose to boost
exploration.

\section{SIMULATION ENVIRONMENT AND CONTROL}
We model the environment in terms of rigid multibodies, frictional contacts, joints, and
motors using the physics engine AGX Dynamics~\cite{Algoryx2017}. For
actuation, we use hinge and linear joints with 1D motors. A 1D motor is a
speed constraint that operates along its remaining degree of freedom by applying a
force/torque to meet a specified target speed.

\subsection{Xt28 forwarder model}
\begin{figure}[ht!]
\centering
\includegraphics[width=0.9\columnwidth]{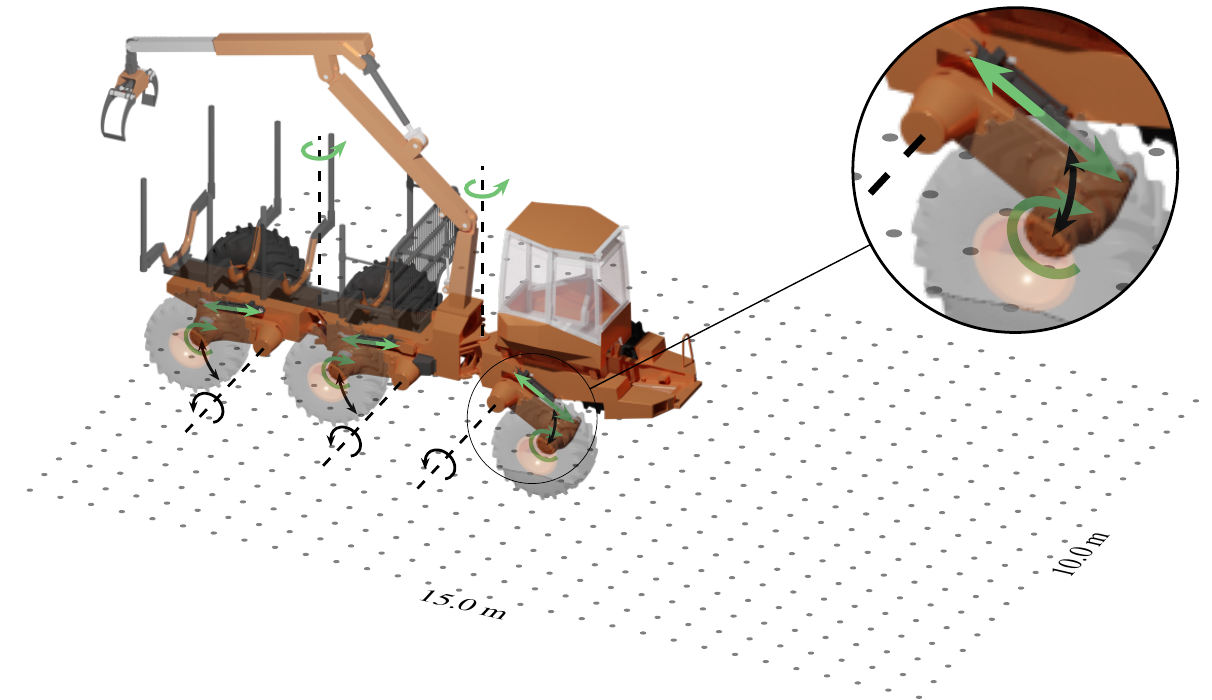}
\caption{The Xt28 model with passive (black arrows) and actuated joints (green
arrows). The frame of reference is located 30~cm behind the cabin. The local height map is represented as a
$15\times10~\mathrm{m}^2$ grid with $30\times20$ resolution.}
\label{fig:xt28-model}
\end{figure}

The Xt28 vehicle, a six-wheeled articulated forwarder, is modelled from a CAD drawing of the actual vehicle as a rigid multibody system with 37 bodies and 14 actuated joints, see Fig.~\ref{fig:xt28-model}.
Hinge motors act at the frame articulation and wheel joints, and linear motors control the suspension arms that are hinged to the chassis.
The wheels are treated as rigid and modelled using spheres due to the computational
benefit in contact detection.

\begin{figure}[ht!]
\centering
\includegraphics[clip, trim=0.0cm 0.0cm 0.0cm 0.0cm, width=0.7\columnwidth]
    {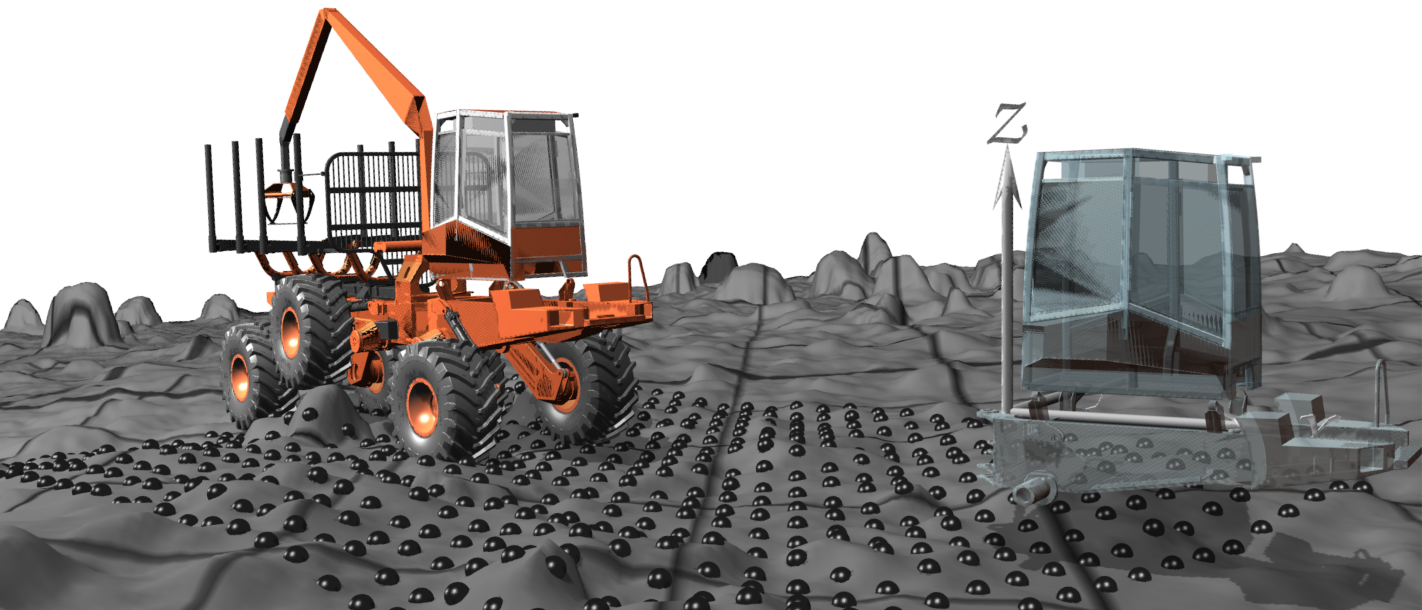}
\caption{Snapshot of the vehicle and target pose given by a position and heading.
The local height map follows the translation and heading of
the vehicle and all heights are taken relative to the height under the reference
frame.}
\label{fig:xt28-on-terrain}
\end{figure}

To have the model state-space agree with the real one, we use realistic masses and
limits on torque and force.  The linear motors have a force limit of 270~kN and the
torque limit at articulation joints and wheel motors are set to 50~kNm and
20~kNm, respectively~\cite{dell2015modelling}. The Xt28 model has a mass of 16
800~kg~\cite{gelin2020concept}, where the centre of mass position of each body
was estimated to match that of the physical vehicle.

\subsection{Controller}
The main goal of the controller is to drive the vehicle to a target pose, given by a position and a heading.
It receives directions to the target $(x, y, \Psi)$, relative to its reference
frame, as well as proprioceptive and additional
exteroceptive information. The proprioceptive information consists of velocities
in the vehicle frame, roll and pitch angles in world coordinates, articulation
frame joint angles, and the piston displacement related to each suspension. It
also receives the longitudinal wheel slip and slip angle. Longitudinal wheel
slip is measured as the difference in forward and surface speed of the wheel normalized by its
forward speed.  The slip angle is the angle between the wheel direction and
the direction it is actually travelling.  Additionally, we observe the load
on each wheel, normalized by vehicle weight. 

The exteroceptive information consists of a local height map of $15\times10~\mathrm{m}^2$
with $30\times20$ grid resolution, which follows the vehicle translation and
heading, see Fig.~\ref{fig:xt28-on-terrain}. In a real world scenario, SLAM, or maps from airborne laser scans of the
terrain together with a GNSS provide similar height maps. The heights are
expressed relative to the reference frame and scaled to be in $[0,
1]$. Together these form a 634-dimensional state representation used by the
controller to select a 14-dimensional action.

For the frame articulation joints and the suspensions, the controller action
specifies target angles and piston positions which are passed to P 
controllers. The P controllers compute the appropriate target speed for each joint motor, operating within their force and torque limits.  The wheel motors
are controlled by setting each motor torque individually. If the angular speed
exceeds $1.5$~rad/s the torque is clamped to not accelerate it
further. Each action is in $[-1, 1]$ and mapped to available joint and torque
ranges.

\subsection{Terrains}
Terrains are constructed from height maps of size $70 \times 70$~m$^2$ and $700
\times 700$ resolution, see Fig. \ref{fig:terrains}. To form a continuous surface, the heights
are interpolated using triangular, piecewise planar, elements into a geometric
mesh. The geometry is assigned to the ground which is represented as a static
rigid body.

There are two different types of terrains. One is procedurally generated
from Perlin noise~\cite{perlin1985image} and semi-ellipsoids to represent
discrete features such as boulders.  The semi-ellipsoids are $[0.5, 3.5]$~m large,
$[0.25, 1.75]$~m tall and the terrain height difference is limited to 5~m. The
procedural terrains are useful for designing training and testing scenarios on
e.g. slopes or terrains with impassable objects at certain locations.

The other type is reconstructions from high-density laser scans, referred to as
\emph{scanned terrains}. Recently 600~Ha of forestry sites were scanned using
600~points/m$^2$ around Sundsvall, Sweden~\cite{LaxsjonDigitalTestsite}. The
dataset is filtered to contain only ground points ($\sim 100$~points/m$^2$) and
converted to a digital elevation model, from which we extract regular
height maps.

\section{LEARNING CONTROL}
To learn a control policy we use PPO because it has proven successful in other
locomotion tasks, e.g.~\cite{xie2020allsteps, peng2018deepmimic}, is easy to
parallelize, and insensitive to hyperparameter settings. The adopted implementation uses
PyTorch~\cite{stable-baselines3} and is based on the original
paper~\cite{schulman2017proximal}. We let the simulation run at 60~Hz and query the
controller at $f_\text{control}=12$~Hz.

\subsection{Network}
As the action space is continuous, a natural choice is to use a diagonal
Gaussian policy, which maps state $s$ to mean actions $\mu_\theta(s)$
represented by a neural network with parameters $\theta$. The variance vector,
$\sigma^2$, is treated as a stand-alone parameter, independent of state. Thus,
the probability of action $a_t$ in state $s_t$ is given by $\pi_\theta(a_t, s_t)
= \mathcal{N}(\mu_\theta, \sigma^2I)$, where $I$ is the identity matrix.

Because part of our inputs are from 2D height maps, we process them separately
with a convolutional neural network. To extract height features, we pass height
maps through two layers with 16 and 32 filter of $3 \times 3$ kernel size,
followed by a fully connected layer with 64 units. We argue that height based
features of importance are similar for the actor and critic and let them share
this part of the network. In the non-shared part, the height map features are
concatenated with the rest of the observations and passed through two fully
connected layers with 128 units each. 
For the actor, the action is produced by a linear output layer of 16 units.
For the critic, the value function is produced by a linear output layer of 1 unit.

\subsection{Reward}
We formulate a reward that encourages steady progress towards the target in an
upright position, without wheel slip, and with limited ground forces, energy
consumption, and damaging tyre sidewall contacts. The net reward takes the form
\begin{equation}\label{eq:reward}
r = r_\mathrm{tar} + 
    r_\mathrm{prog} r_\mathrm{roll} r_\mathrm{speed} r_\mathrm{forces}
    \times \frac{r_\mathrm{head} + r_{\mathrm{slip}\parallel} + r_{\mathrm{slip}\perp}}{3}
    + r_\mathrm{energy} + r_\mathrm{side},
\end{equation}
where the terms are explained below.
The main goal of the controller is met when the vehicle is closer than
0.3~m and $9^{\circ}$ relative to the target position and heading. We
define the target bonus as
\begin{equation}
    r_\mathrm{tar} = k_\mathrm{tar} \mathds{1}(\Psi, d_t),
\end{equation}
where $k_\mathrm{tar}$ is a constant set to 5 per cent of the maximum,
undiscounted return and $\mathds{1}$ is the indicator function which evaluates
to 1 at the target and 0 otherwise.

The target reward yields a sparse signal unlikely to be discovered in early
stages of training. As guidance we provide a dense reward which
reflects the progress toward the target as
\begin{equation}
    r_\mathrm{prog} = (d_{t - 1} - d_t)f_\mathrm{control},
\end{equation}
where $d_t$, $d_{t-1}$ is the current and previous distance from the vehicle to
the target projected to the horizontal plane.
We reason that heading alignment is increasingly important as the vehicle 
approaches the target and introduce it as a reward multiplier 
\begin{equation}
r_\mathrm{head} = 
    \exp \left[-\frac{1}{2}
            \left( \frac{\Psi}{d_t / k_\mathrm{d}} \right)^2 
         \right],
\end{equation}
where the constant $k_\mathrm{d}=5$~m is tuned with the turning 
radius of the vehicle.

In the reward shaping process we observed that a reward $r = r_\mathrm{tar} + 
r_\mathrm{prog} r_\mathrm{head}$ is essential for learning to reach the
target quickly, but does not promote efficient, safe and environmental friendly driving.
Therefore, we introduce a set of additional reward multipliers with range $[0, 1]$.
To avoid risk of overturn we define the roll reward as
\begin{equation}
r_\mathrm{roll} = 
    \exp \left[-\frac{1}{2}
    \left( \frac{\phi}{k_\mathrm{\phi}} \right)^2
    \right],
\end{equation}
for roll angle $|\phi| > 5^{\circ}$ and 1 else, where we use $k_\mathrm{\phi} =
\pi / 16$.
To encourage limited vehicle speeds, we use 
\begin{equation}
r_\mathrm{speed} = \min(1, \exp[k_\mathrm{speed} (v_\mathrm{lim} - |v|)]),
\end{equation}
where $v_\mathrm{lim} = 0.8$~m/s, and $k_\mathrm{speed}=2$ is a constant manually tuned to control the rate of
reward decay for speeds above $v_\mathrm{lim}$.

To limit ground pressure we consider the standard deviation of normalized ground
forces, $\sigma_\mathrm{forces}$. 
Ground pressure is at its lowest in case of an even distribution over the 6 wheels.
Each wheel then carries 1/6 of the vehicle weight, and $\sigma_\mathrm{forces} = 0$.
We promote even weight distribution through
\begin{equation}
r_\mathrm{forces} =
    \exp \left[-\frac{1}{2}
    \left( \frac{\sigma_\mathrm{forces}}{k_\mathrm{forces}} \right)^2
    \right],
\end{equation}
where $k_\mathrm{forces} = 0.1~\mathrm{N}$.

Reaching the target is not considered a success with excessive slip during the
episode. Therefore we include two terms related to longitudinal slip
$\lambda$ and slip angle $\alpha$ as
\begin{align}
r_{\mathrm{slip}\parallel} &= \prod_i^{n_{wheels}}
    \exp \left[-\frac{1}{2}
    \left( \frac{\lambda_i}{k_\mathrm{\lambda}} \right)^2 
    \right],
    & k_\mathrm{\lambda} & = 0.3\\
r_{\mathrm{slip}\perp} &= 
    \prod_i^{n_\mathrm{wheels}} 0.5 \cos(k_\mathrm{\alpha} \alpha_i) + 0.5,
    & k_\mathrm{\alpha} & = 6,
\end{align}
where $\alpha_i$ is clipped at $\pm ~ \pi / k_\mathrm{\alpha}$ such that any
slip angle outside that range yields zero reward.
The slip rewards are constructed as products to induce well behaved wheel motions 
for all wheels simultaneously. The slip and heading terms are mutually conflicting objectives. Therefore we sum them to a single multiplier, as seen in \ref{eq:reward}.

To promote smooth, efficient motions, energy consumption is included as
\begin{equation}
    r_\mathrm{energy} =
        k_\mathrm{energy} \frac{W_\mathrm{joints}}{W_\mathrm{max}},
\end{equation}
where $W_\mathrm{joints}$ is the total work carried out by all actuated joints over the previous action step,
${W_\mathrm{max}}$ is its upper bound, and $k_\mathrm{energy} = -1$.

Damage to tyre sidewalls are penalized through the number of sidewall contacts
$n_\mathrm{contacts}$ as
\begin{equation}
    r_\mathrm{side} = k_\mathrm{sw} n_\mathrm{contacts},
\end{equation}
where $k_\mathrm{sw} = -0.2$. We found this reward term necessary to avoid use of
the sides of the tyres for traction. A contact is classified as being on the
sidewall if the angle between the contact point in the wheel frame and
the rotational axis is less than 60$^\circ$. 

A nice feature of the reward in (\ref{eq:reward}) is that the maximum undiscounted
return is easily calculated as the initial distance to the target,
times the control frequency, plus the target reward. Although, the maximum is not
attainable in practice, is serves as good reference for designing a curriculum and
evaluating policy performance.

\section{Training and evaluation}
During training, an episode starts with the vehicle being deployed on the
terrain at random position, $x_0, y_0~[\mathrm{m}] \in [-1, 1]$, and heading $\psi_0 \in [0,2\pi]$. We
let the vehicle drop to the ground and settle. To get natural variations of
initial vehicle configurations we apply a simple controller to the suspensions
during a simulated time period of 1~s.

\begin{figure}[htb!]
\centering
\includegraphics[width=0.6\columnwidth]{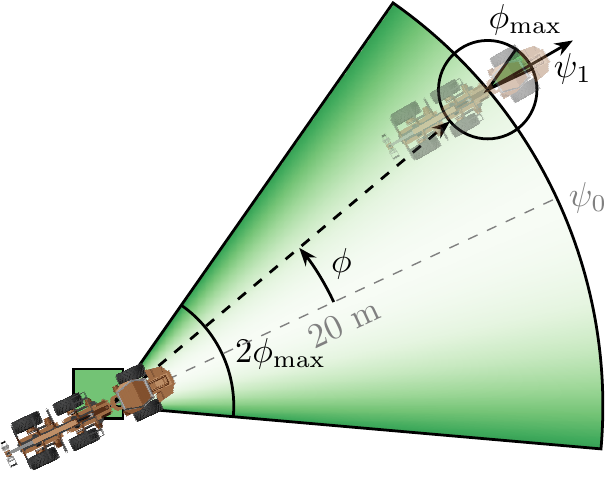}
\caption{Target generation. The vehicle is initialised in the green square with
random heading. The target is then placed a distance 20~m away along a limited
arc with heading $\psi_1$.}
\label{fig:target_generation}
\end{figure}

To enable curriculum with altered target difficulty, a \emph{target heading parameter} $\phi_{\text{max}}$ is defined, affecting both target placement and heading.
The target is placed a distance $r_0 = 20$ m away, within an angle $\phi \in [-\phi_\text{max},  \phi_\text{max}]$ relative to the vehicle heading, see
Fig.~\ref{fig:target_generation}. To put emphasis on learning stearing, the
target position along this arc is sampled from a quadratic distribution,
increasing the probability toward the edges. The target heading is then sampled from
a uniform distribution, $\psi_1 \in [-\phi_{\text{max}}/2, \phi_{\text{max}}/2]$
relative to $\phi$, i.e. the angle \emph{to} the target.

A training episode runs until the target is reached, or terminated after
400 or 500 steps, depending on the curriculum. Early termination occurs if the
vehicle moves beyond the target, if it reaches a roll beyond $25^\circ$, or if a
terminal contact is detected. A terminal contact is when any part of the chassis
comes in contact with the terrain. 

Training is done on 10 parallel environments on a cluster with 28 cores, where
each environment uses a different terrain. After every 25k steps the controller
is evaluated in a separate environment on a terrain not used in training, with
deterministic initial vehicle positions, target placements, and action selection
based on the latest policy.

\subsection{Curriculum} \label{sec:curriculum}
In our experience, a curriculum is essential for the controller to reach its full potential.
Our goal is to form a curriculum such that there is a solid foundation in basic driving skills after the first lesson, e.g. acceleration, turning, and speed control.
The purpose of the following lessons is to specialize driving skills towards preference.
To emulate natural forest environments, we focus on boulder-like obstacle avoidance, unevenness, and slopes.

\begin{figure}[htb!]
\centering
\includegraphics[width=0.8\columnwidth]{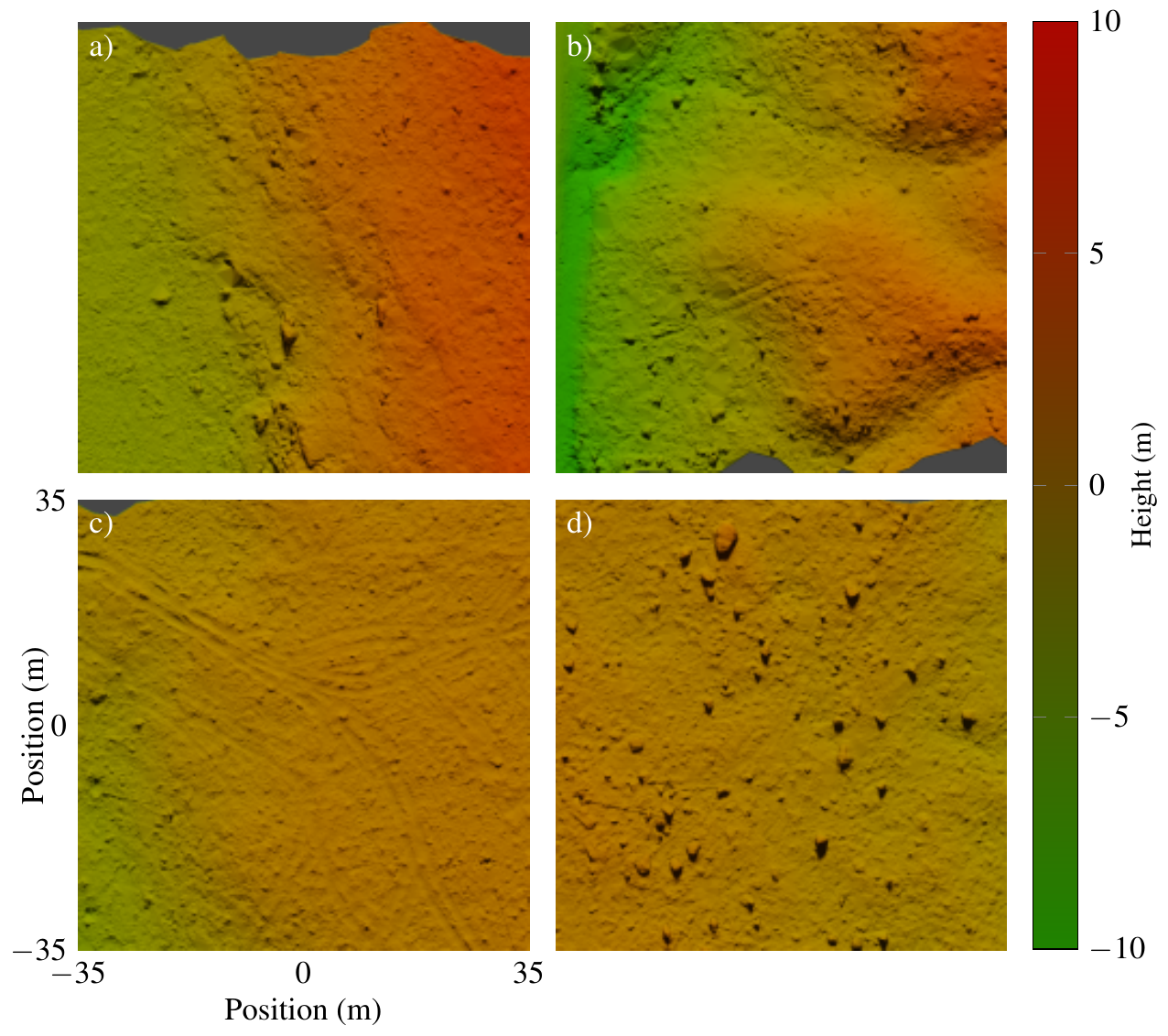}
\caption{Patches from scanned terrains. a) and b) are two examples used for training, c) is used in domain sensitivity experiments, and d) in obstacle perception. The images are rendered with terrain colour according to height.}
\label{fig:terrains}
\end{figure}

Our approach is to use a fixed order boundary curriculum~\cite{xie2020allsteps}
for the terrain and target placements, where the learning process is divided
into four lessons with increasing difficulty according to our intuition. In the
simplest, initial lesson, the terrain is level with Perlin noise to mimic
features of natural terrain. To put emphasise on sharp turns we set the target
heading parameter to $\phi_\mathrm{max} = \pi / 3$ already in the first lesson.
The second lesson focuses on learning height map features to avoid impassable
objects. We use the same terrain base with Perlin noise but add 8
semi-ellipsoids placed randomly between the initial vehicle position and the target.
To both avoid obstacles and reach the target is challenging, so we simplified
the task by setting $\phi_\mathrm{max} = \pi / 9$. The third level uses a similar
setting with tougher Perlin noise to form a hilly/slopy terrain, but with only 6
impassable semi-ellipsoids and 6 smaller ones. In the final level, the
controller practices driving on scanned terrains with $\phi_\mathrm{max} = \pi / 3$
and 500 max steps. We chose terrain patches that appeared trafficable,
yet challenging with steep slopes, boulders, and ditches, see
Fig.~\ref{fig:terrains}.

\subsection{Hyperparameters}
For the PPO related hyperparameters we use a horizon of 1280, minibatch size of
800, and 10 epochs. We use the Adam optimizer with a gradually decreased step size between lessons.
A step size of $25\times 10^{-5}$ is used in the first, $10\times 10^{-5}$ in the second and third, and $1\times 10^{-5}$ in the forth lesson respectively.
The discount is $\gamma=0.99$ and the GAE parameter
$\lambda = 0.95$. The value function and policy both have clipping range $0.2$.
The value function coefficient for the loss calculation is $0.5$ and the entropy
coefficient $0.01$.

\section{RESULTS AND DISCUSSION}
\begin{figure}[hbt!]
\centering
\includegraphics[width=0.9\columnwidth]{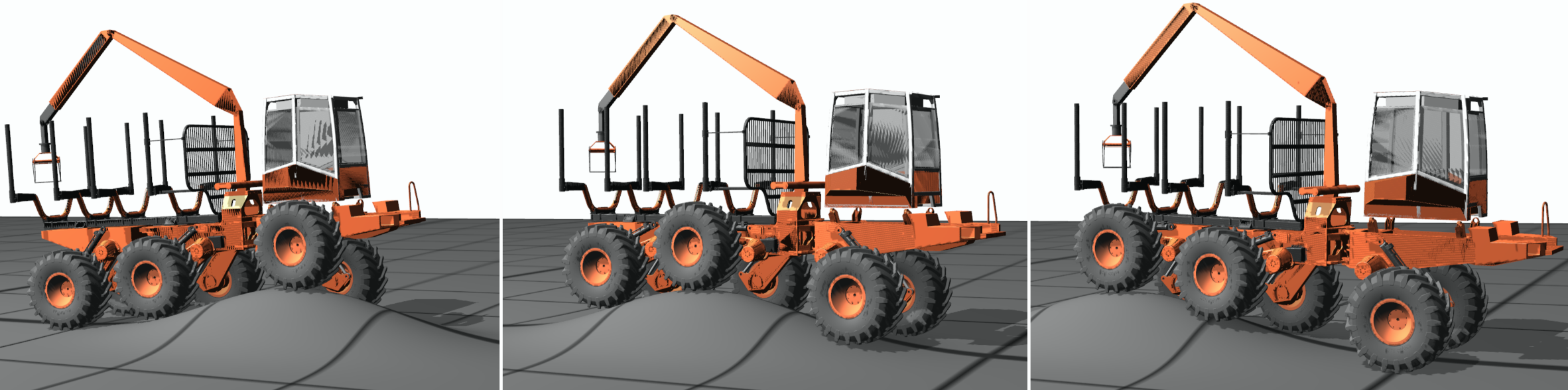}
\caption{Sequential snapshots of the vehicle traversing a 1~m tall gaussian bump,
avoiding chassis roll and wheel slip.}
\label{fig:results:gaussian}
\end{figure}

We present a controller that shows smooth progression towards the target
while adapting to terrain irregularities.  When turning, torques are
adjusted so that the outer wheels rotate faster than the inner, thereby moving
with limited slip and effort. The suspensions are used conservatively and kept
in fixed position unless the vehicle is challenged by slopes or unevenness in
the terrain. When faced with a Gaussian bump of 1~m height, the controller makes
intelligent use of the suspensions for levelling and ground compliance, as shown
in Fig.~\ref{fig:results:gaussian}. The maximum slip is 1.5\% and the average
slip per wheel is only 0.15\%. To see highlights of the learnt driving skills on
a number of different terrains we refer to the supplementary video.

\begin{figure}[hbt!]
\centering
\includegraphics[width=0.6\columnwidth]{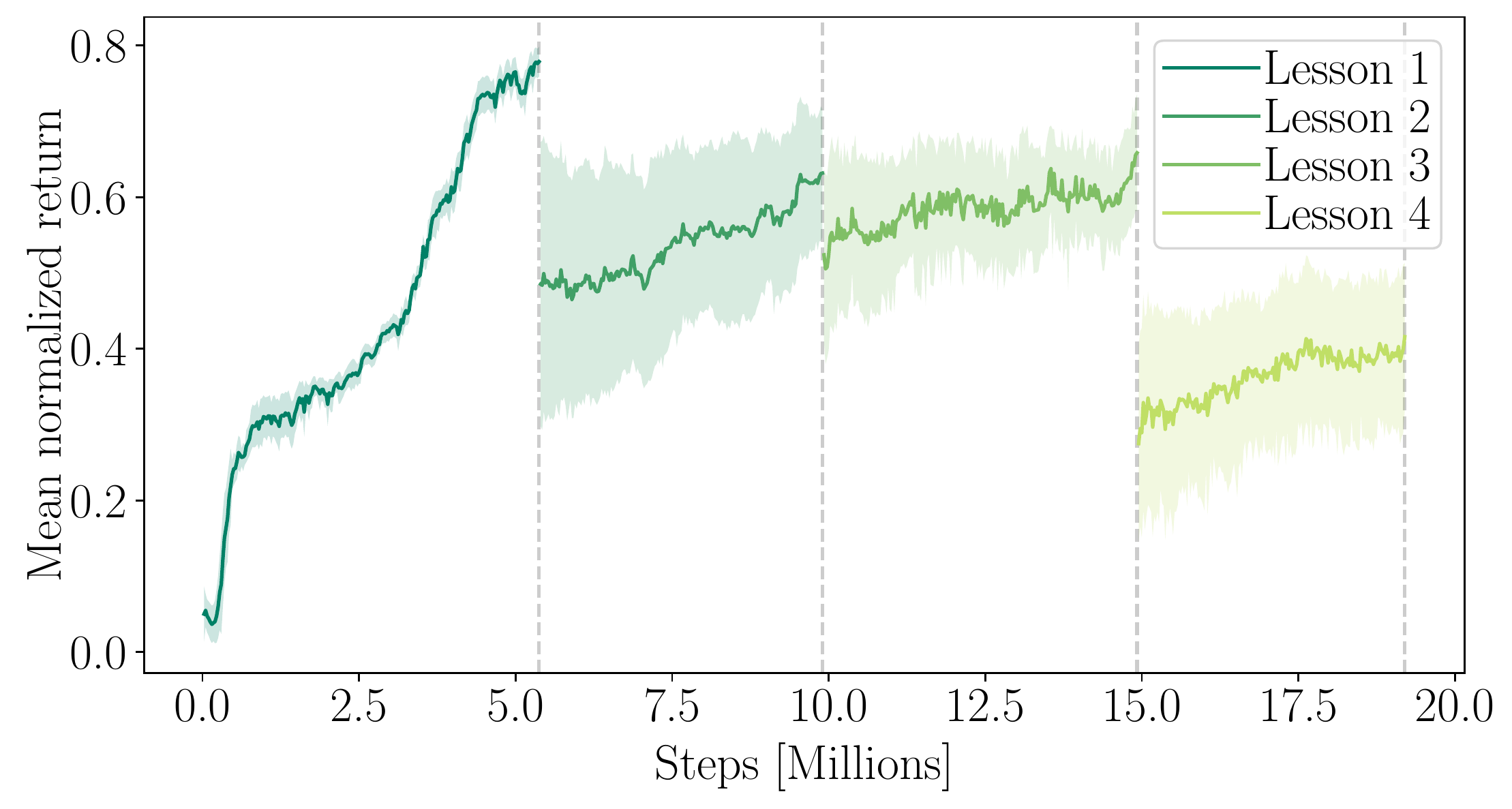}
\caption{Learning curves over four consecutive curriculum lessons with increasing
difficulty. The controller was evaluated every 25~k steps over 20 episodes with
deterministic action selection.}
\label{fig:results:learning}
\end{figure}

Training is done according to the curriculum in Section~\ref{sec:curriculum}, where the
best policy in the preceding lesson is used as starting point for the next, see Fig.~\ref{fig:results:learning}. In total, the controller is trained
for 19.22~M steps and 108~h CPU hours. Learning is rapid during the first
lesson except during a plateau. We found that penalizing energy consumption was
key to develop strategies to limit speed and keep progressing, but it also 
eliminated jerky and unnecessary movements.


\subsection{Sloped terrains}\label{sec:sloped-terrains}
\begin{figure}[hbt!]
\centering
\includegraphics[width=0.98\columnwidth]{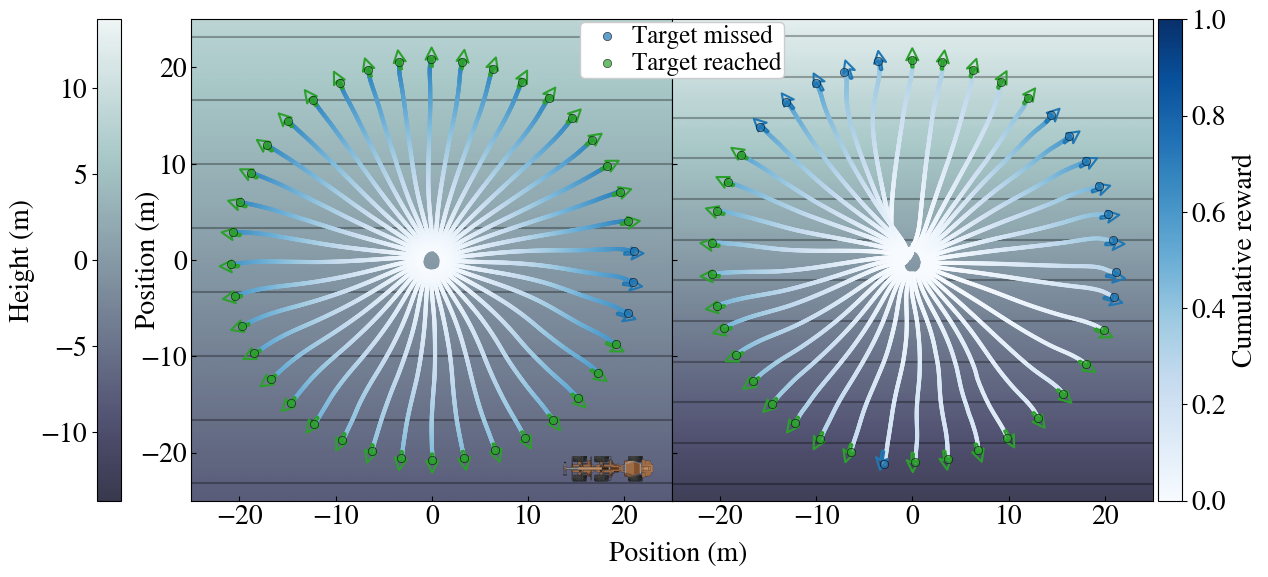}
\caption{Comparison of controller performance on two terrains with 18 and
27$^\circ$ incline. The arrows show target placements. A target is reached when the vehicle is closer than 0.3~m and 9$^\circ$ relative to the target position and heading. The vehicle is true to scale.}
\label{fig:results:sloped}
\end{figure}

The controller shows the ability to traverse steep slopes and uses different strategies
depending on the slope direction. We use two perfectly even
terrains with 18$^\circ$ and 27$^\circ$ incline, and place the vehicle
around the centre, with equally spaced heading in 40 directions following a full
rotation, see Fig.~\ref{fig:results:sloped}.
The success rates are 92.5\% and 65\%  with undiscounted mean
normalized return $0.64 \pm 0.09$ and $0.40 \pm 0.13$ for the 18$^\circ$ and
27$^\circ$ terrains, respectively.
As reference, the terrains are rated as 4/5 and 5/5 in difficulty according to the terrain classification system for forestry work in nordic countries~\cite{berg1982terrangtypschema}.
On side slope, the controller utilizes one of the claimed benefits of the Xt28 and adjusts the suspensions to
shift the centre of mass and maintain an upright position in an attempt to minimize
ground forces, wheel slip, and roll. We note that the maximum side slope which
allows for complete levelling is 27.5$^\circ$ due to the range limits of the suspensions.
Even so, the mean rolls are $1.93 \pm 0.94^\circ$ and $3.83 \pm 2.59^\circ$,
respectively, including the unfavourable initial configurations.
Although the success rate is
not as high for the steeper terrain, there are no complete failures, and the
missed targets are typically due to side slip. Curiously, it is more demanding to drive downhill than uphill. The loss in reward is mainly due to the inability to
maintain speed below the upper limit.

\subsection{Obstacle perception}
\begin{figure}[hbt!]
\centering
\includegraphics[width=0.99\columnwidth]{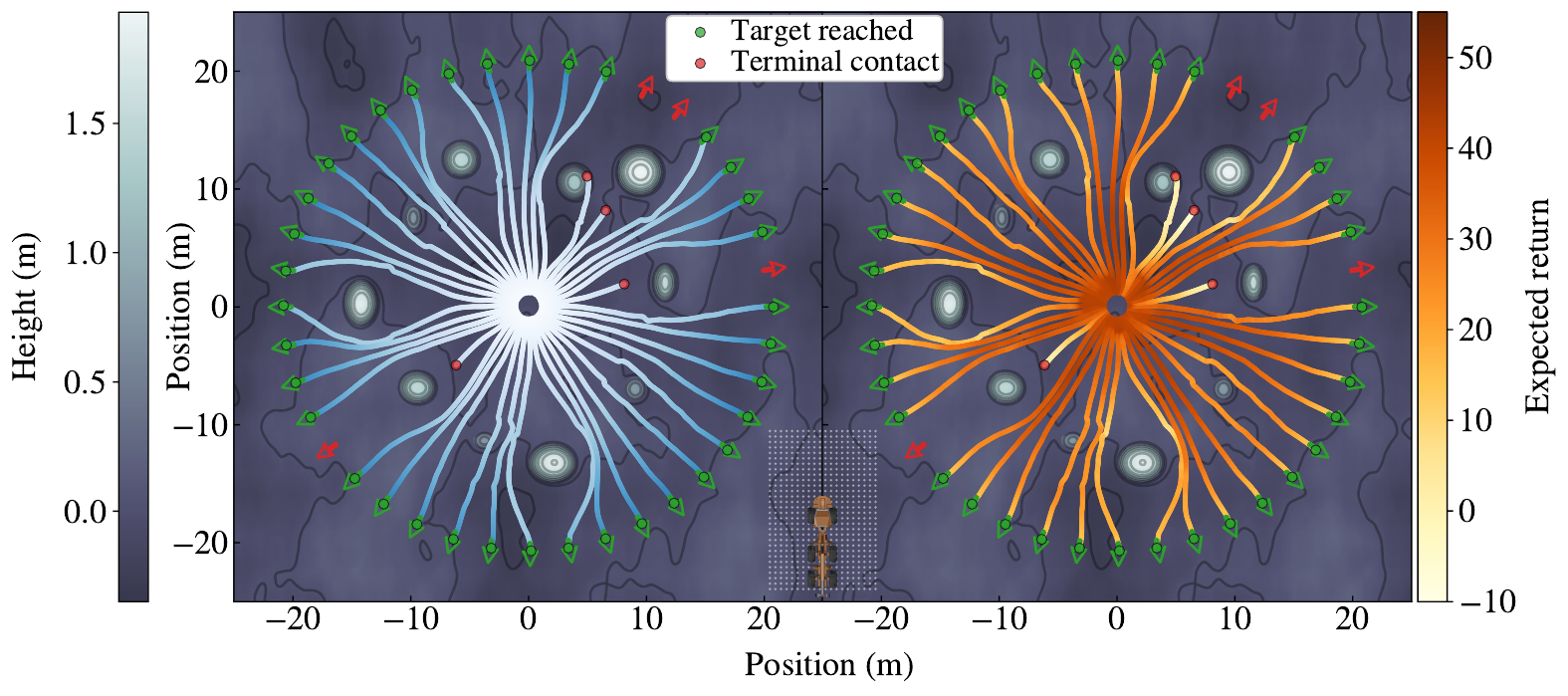}
\caption{Motion trajectories on procedurally generated terrain with
semi-ellipsoids representing boulders. The trajectories are coloured by
normalized cumulative reward in $[0,1]$ (left) and the learnt value function
estimates (right). The controller displays the ability to perceive by driving
around impassable objects and over smaller. The vehicle and local height map is
true to scale.}
\label{fig:results:boulders}
\end{figure}

If faced with objects of different sizes, the controller shows an ability to
distinguish between passable and impassable ones and places the wheels to avoid
sidewall contacts. To see the strategies we test the controller on a terrain
similar to those with semi ellipsoids used in training, see
Fig.~\ref{fig:results:boulders}. Targets and initial vehicle positions are the
same as for the sloped terrains, resulting in a 90\% success rate and
undiscounted mean normalized return of $0.62 \pm 0.15$ over 40 episodes.
Impassable objects that appear within the range of the local height map are well
reflected in the value function estimates, far before reaching the problematic
location. States with impassable objects straight ahead are expected to result
in poor performance unless easy to circumvent, at which the trajectory is
planned by taking out turns enough to avoid contact and reach the target
placement. Smaller objects are easily overcome without significant loss in reward due to efficient use of the suspensions. Because some episodes are practically
impossible and require going in reverse, a driving skill not practised during
training, we cannot expect full success. The four episodes with terminal
chassis-ground contacts occur when the vehicle is directly facing large objects
and is unable to choose which way to turn.

\begin{figure}[hbt!]
\centering
\includegraphics[width=0.99\columnwidth]{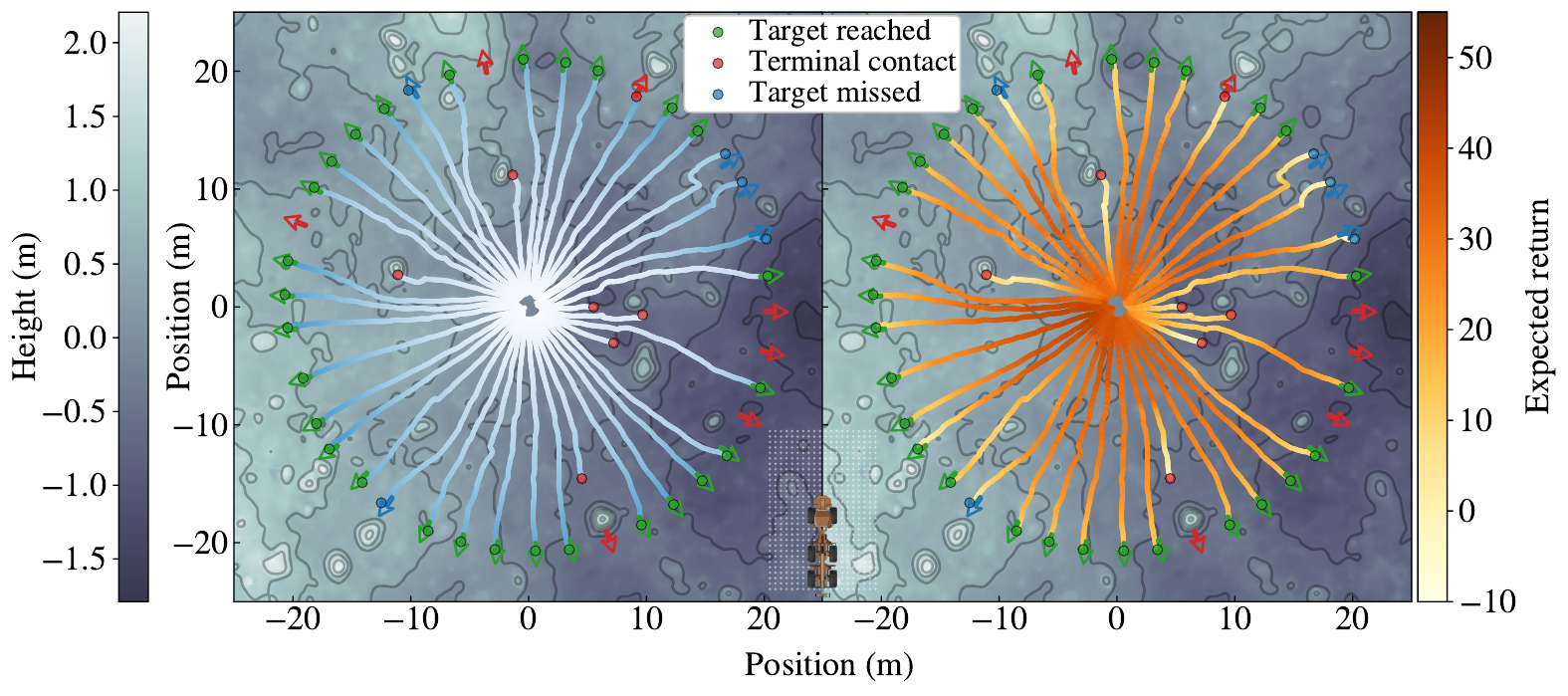}
\caption{Obstacle perception on a scanned rough terrain. The trajectories are
coloured by normalized cumulative reward (left) and value function estimates
(right).}
\label{fig:results:boulders_LAS}
\end{figure}

To test if the learnt skills generalize to natural environments we repeat our
previous experiment on a terrain patch extracted from the real data set. The
selected area (Fig.~\ref{fig:terrains}d) contains the highest density of
large boulders ($>1$~m tall) from the 600~Ha test site and poses a severe
challenge. The target is reached 70\% of episodes with a mean normalized return
of $0.48 \pm 0.14$, see Fig.~\ref{fig:results:boulders_LAS}. The results are
similar to the artificial terrains, where the controller surpasses smaller
boulders, circumvents others, and the majority of unsuccessful episodes is due
to chassis-ground collisions. We note that most terminal contacts occur when the
target is in the vicinity of a large boulder or when several boulders obstruct
the passage, e.g. east in Fig.~\ref{fig:results:boulders_LAS}. Without a clear
passage, the expected return is immediately small, indicating that the
controller recognizes when put to a task it cannot successfully complete. To
further study the value function is valuable if we want to enhance obstacle
perception. However, when it comes to obstacle avoidance, it is not clear if the
responsibility should lie completely in a low level controller or one at higher level doing path planning.


\subsection{Smart control on real forest terrain}
\begin{figure}[hbt!]
\centering
\includegraphics[width=0.7\columnwidth]{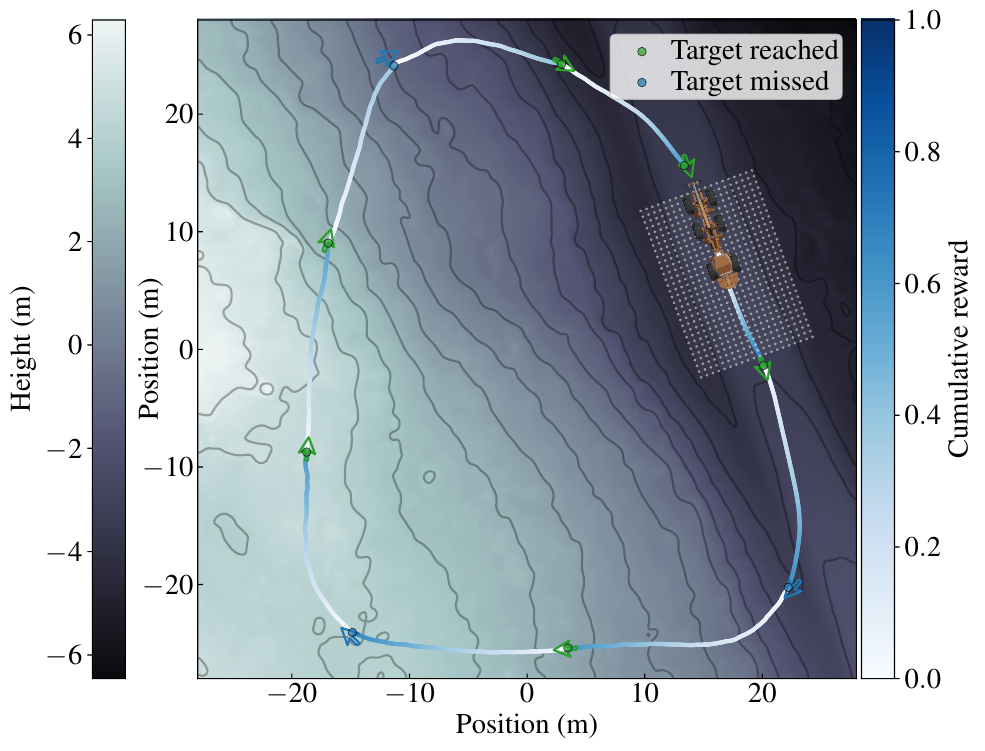}
\caption{Top view of vehicle trajectories following
a sequence of waypoints placed on a reconstruction of real terrain from high-density
laser scans. The vehicle starts and ends at a primary road along a route similar
to a real world forestry scenario.}
\label{fig:results:LAS-2d}
\end{figure}

\begin{figure}[hbt!]
\centering
\includegraphics[width=0.6\columnwidth]{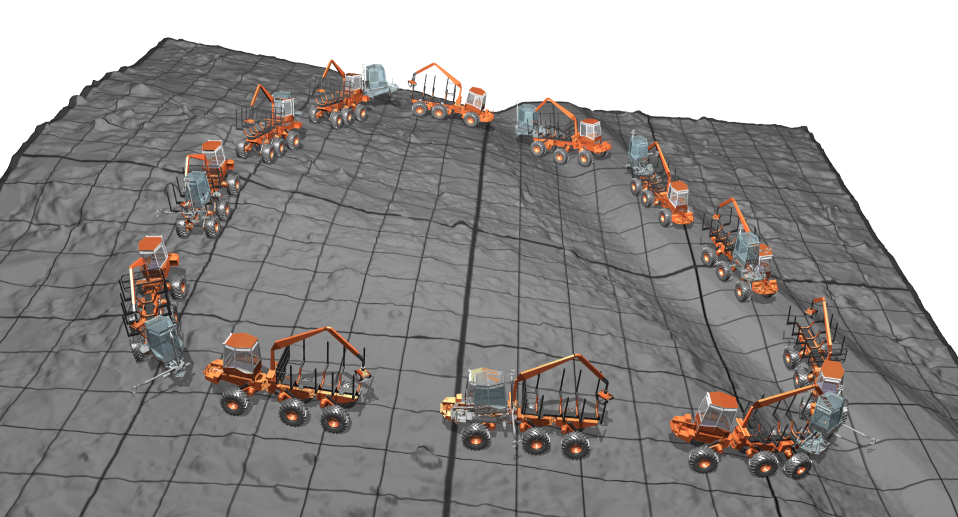}
\caption{3D rendering of the vehicle and waypoints.}
\label{fig:results:LAS-3d}
\end{figure}

To simulate the use of the controller in a purposeful forestry application we test its
driving skills on scanned terrains. We emulate a higher level planner and
manually place a sequence of targets, or waypoints, starting and ending at a
primary road to complete a full cycle, see Figs.~\ref{fig:results:LAS-2d} and
\ref{fig:results:LAS-3d}. The terrain has a mean slope of $12^\circ$, a deep
ditch alongside the road, and enough roughness to serve as a challenging
test.

Despite being a difficult route on demanding terrain 6 out of 9 waypoints are reached,
where the misses are small and do not affect the higher level goal of completing
the route. The controller displays an ability to cross ditches, a challenging
real world scenario, and handles target placements not seen in training with
ease. The mean normalized return is $0.60 \pm 0.12$ where, as discussed with
sloped terrains, the vast majority of lost reward comes from driving too fast
downhill. Still, there is no tendency towards unsafe traversal and we note that
the top speed was no more than 0.37~m/s above limit.

\subsection{Domain sensitivity}
\begin{figure}[ht!]
\centerline{\subfigure[]
    {\includegraphics[width=0.5\columnwidth]{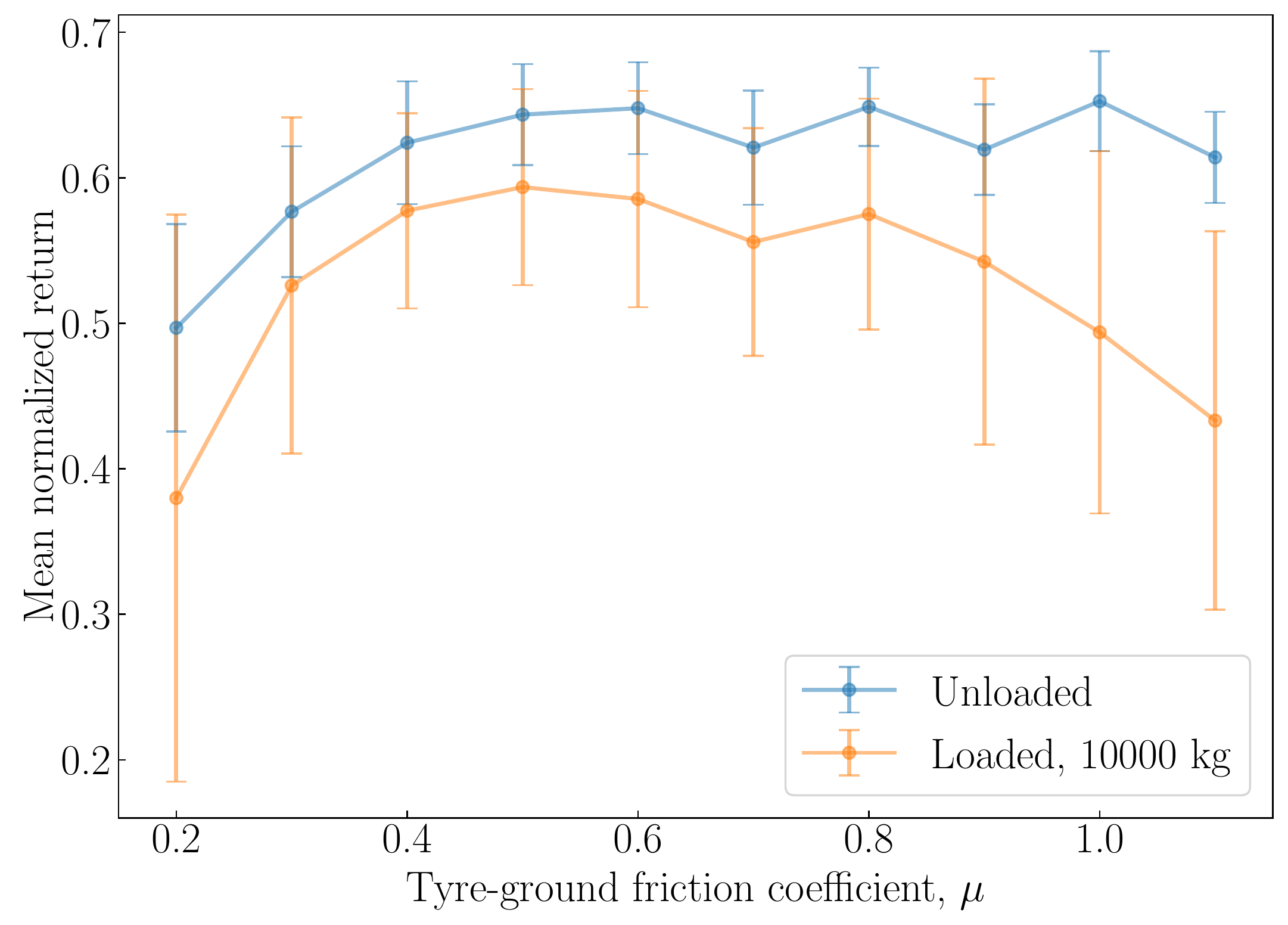}
    \label{fig:results:LAS-domain-sensitivity-a}}
\hfil
\subfigure[]
    {\includegraphics[width=0.5\columnwidth]{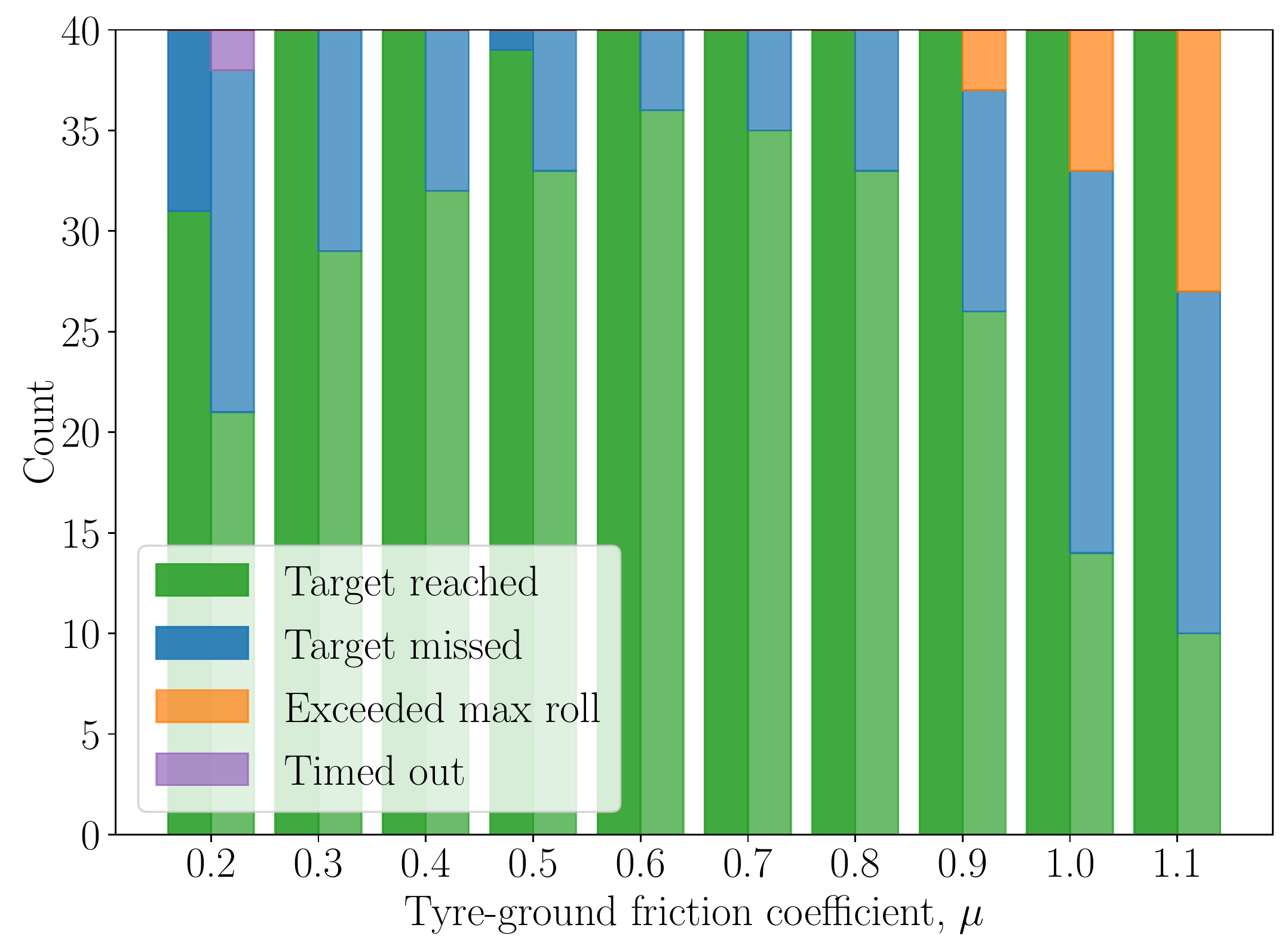}
    \label{fig:results:LAS-domain-sensitivity-b}}}
\caption{a) Undiscounted mean normalized return over 40 episodes as a function
of tyre-ground friction coefficient, $\mu$, where the error bars show one
standard deviation. The vehicle is either unloaded or carries 10000~kg. b)
Episode termination. The left bar in each pair corresponds to an unloaded
vehicle and the right, slightly brighter, to one with 10000~kg load.}
\label{fig:results:LAS-domain-sensitivity}
\end{figure}

The controller is insensitive to variations in ground-terrain friction
coefficient $\mu$, and able to adapt to load cases not seen during training. In
natural environments, surface friction varies over space and time, while
variable load is relevant in any transport application, e.g. forestry,
agriculture. We chose a typical forestry site from the real dataset
(Fig.~\ref{fig:terrains}c) and let $\mu \in \{0.2, 0.3, \ldots, 1.1 \}$ for
two vehicle load cases: one with nominal weight and another where a static
10000~kg load (60\% weight increase) is placed on the load bunk. The targets are placed
20~m away with random heading $[-\pi/3, \pi/3]$, relative to the vehicle
starting position. For each of the 20 cases we simulate 40 episodes and compute
the undiscounted mean normalized return and standard deviation, see
Fig.~\ref{fig:results:LAS-domain-sensitivity-a}.  

As expected, the controller performs at its best around the settings used for
training, i.e., unloaded with $\mu = 0.7$, and equally well for higher friction.
Performance is not significantly affected until $\mu$ drops below 0.4, which
roughly corresponds to the average sliding friction between tyres and wet earth
roads~\cite{wong:2001:tgv}. From
Fig.~\ref{fig:results:LAS-domain-sensitivity-b}, it is clear that the target
is frequently reached at $\mu = 0.3$, but more seldom for $\mu = 0.2$. The
loaded case shows similar behaviour but with ~10\% lower episodic return. To
some degree this is due to the higher energy consumption with the increase in
weight, but Fig.~\ref{fig:results:LAS-domain-sensitivity-b} shows that in
10-20\% of the cases, the heavier vehicle fails to reach the target. Notably,
performance drops for friction above $0.8$, where a fair portion of episodes
terminate due to maximum roll being exceeded. The high friction and load resists
turning at moderate speed and the controller compensates by tilting to increase
traction on the outer wheels. With no experience in similar states, it proceeds
until failure occurs.

\begin{figure}[hbt!]
\centering
\includegraphics[width=0.6\columnwidth]{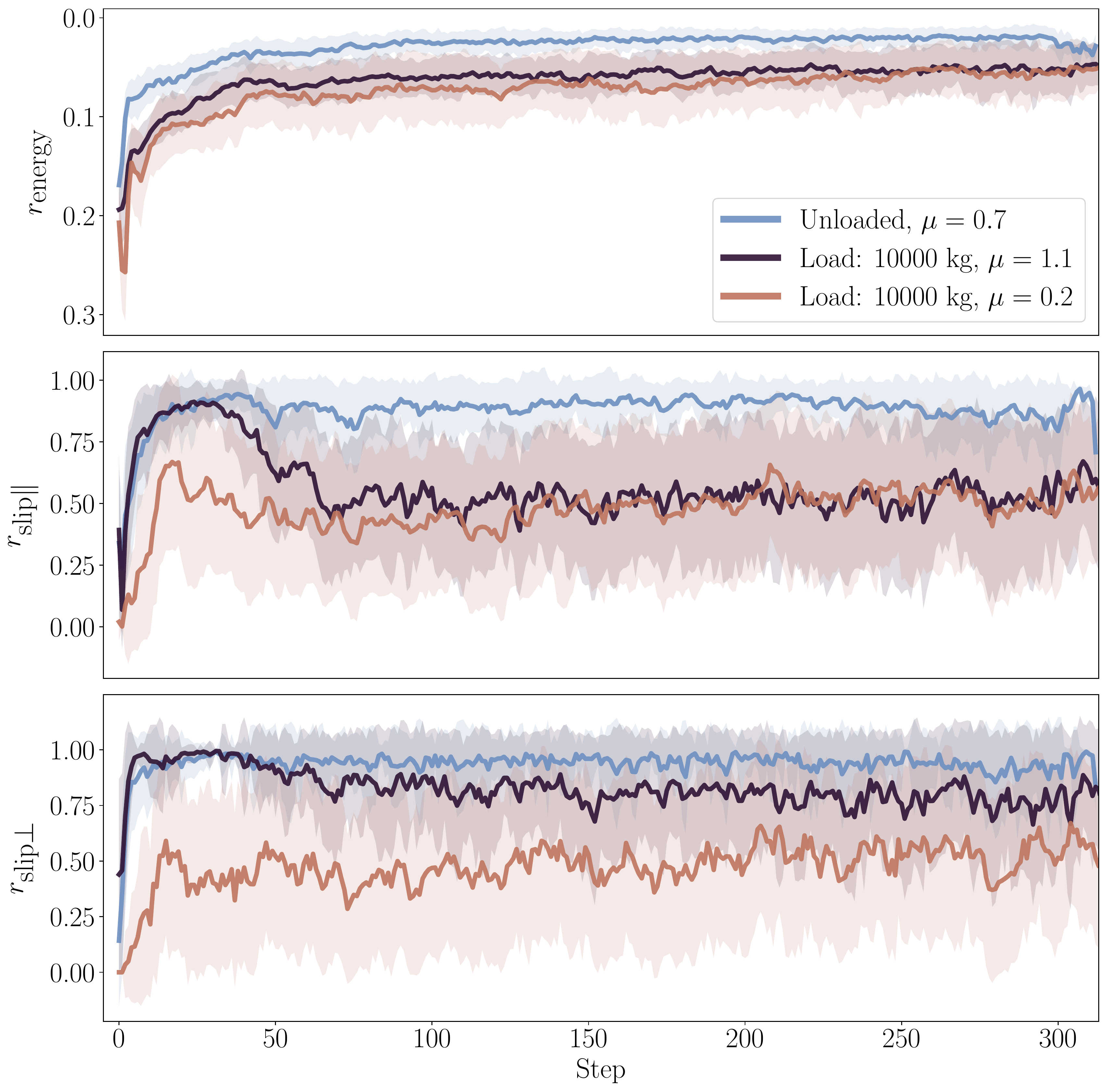}
\caption{Mean reward contributions and standard deviation over 40 episodes for different friction and vehicle load. Number of steps was truncated at the shortest episode.}
\label{fig:results:LAS-domain-rewards}
\end{figure}

To further understand the effect of different vehicle load and ground-tyre
friction on performance we look at individual reward contributions.
Fig.~\ref{fig:results:LAS-domain-rewards} shows $r_\mathrm{energy},
r_{\mathrm{slip}\parallel}$, and $r_{\mathrm{slip}\perp}$ for the two cases with
lowest mean return, and training settings. Not surprisingly, low friction and
added load leads to an increase in energy consumptions and slip. We observe that a
loaded vehicle in high friction setting drives with significantly less slip
compared to low friction, but similar side slip except in the first quarter of
episodes. This again is due to the resistance in turning, and also the
difficulties to control the frame articulation.  

\section{CONCLUSIONS}
We conclude that deep RL is more than capable of learning control for rough
terrain vehicles with continuous, high dimensional, observation, and action
space. We have presented a controller that perceives, plans, and individually
controls six suspensions, six wheels, and two frame articulation joints, without
the use of frame stacking or recurrent networks as memory support. The
controller relies on a local height map to perceive which obstacles to
circumvent, how to handle steep slopes, etc., and then couples its perception
with proprioceptive features to efficiently traverse rough terrain. The
traversal is done with minimal slip, roll, and energy consumption, to reach a
target placement. The controller is robust to friction between tyre and ground,
as long as it does not fall below a critical value. It is more sensitive to
changes in the vehicle weight, which poses a problem when collecting and
transporting heavy objects.
We suggest that deep RL will be a future cornerstone for control of vehicles with high
dimensional state space, especially in environments where it is easier to react
to the dynamics than predict them with sufficient accuracy.

\section*{ACKNOWLEDGEMENTS}
This work has in part been supported by Mistra Digital Forest (Grant DIA
          2017/14 6) and Algoryx Simulation AB. The simulations were performed on
          resources provided by the Swedish National Infrastructure for Computing (SNIC
          410 dnr 2021/5-234) at High Performance Computing Center North (HPC2N).

\addtolength{\textheight}{-12cm}   

\def\cprime{$'$}

\end{document}